\documentclass{article}
\usepackage{lex_sem_rel}

\usepackage{times}
\usepackage{soul}
\usepackage{url}
\usepackage[hidelinks]{hyperref}
\usepackage[utf8]{inputenc}
\usepackage[small]{caption}
\usepackage{graphicx}
\usepackage{amsmath}
\usepackage{booktabs}
\usepackage{amssymb}
\usepackage{multirow}
\usepackage{subcaption}
\title{Assessing the Lexico-Semantic Relational Knowledge Captured by Word and Concept Embeddings}

\author{
Ronald Denaux \and Jose Manuel Gomez-Perez \\
\affiliations
Expert System Cogito Labs\\
\emails
\{rdenaux,jmgomez\}@expertsystem.com
}

\begin{document}

\maketitle

\begin{abstract}
Deep learning currently dominates the benchmarks for various NLP tasks and, at the basis of such systems, words are frequently represented as embeddings --vectors in a low dimensional space-- learned from large text corpora and various algorithms have been proposed to learn both word and concept embeddings. One of the claimed benefits of such embeddings is that they capture knowledge about semantic relations. Such embeddings are most often evaluated through tasks such as predicting human-rated similarity and analogy which only test a few, often ill-defined, relations. In this paper, we propose a method for (i) reliably generating word and concept pair datasets for a wide number of relations by using a knowledge graph and (ii) evaluating to what extent pre-trained embeddings capture those relations. We evaluate the approach against a proprietary and a public knowledge graph and analyze the results, showing which lexico-semantic relational knowledge is captured by current embedding learning approaches.
\end{abstract}
\section{Introduction and Related Work}
Most of the recent interest in the area of \textbf{word embeddings} was triggered by the Word2Vec algorithm proposed in \cite{Mikolov2013}, which provided an efficient way to learn word embeddings by predicting words based on their context and using negative sampling. Word embeddings have become the usual input to natural language processing (NLP) tasks, but also tasks for which previously knowledge graphs were being used. Applications range from text classification \cite{Kim2014} to machine translation \cite{Kalchbrenner2013,Sutskever,Cho2014,Bahdanau2014}, question answering \cite{Khot,Seo,Parikh} and knowledge graph construction and completion\cite{fu2014semHierarchiesViaEmbeddings}. However, despite recent efforts \cite{Li,Garcia2018,Zeiler2014}, the nature and extent of the knowledge captured by such embeddings and how they contribute to accomplish the goal in question is still hard to interpret.

Embeddings have shown the ability to learn relations between words. However, most benchmarks \cite{Schnabel2015} focus on a specific family of relations involving relations similarity and analogy. It is unclear whether such relations simply happen to be well aligned with the statistic analysis involved in the computation of the embeddings, or whether the embeddings are capable of capturing a wider range of relational knowledge. Indeed, standard benchmarks show evidence that word embeddings may capture more types of relations. It has been shown that algorithms like FastText \cite{Bojanowski2017}, GloVe \cite{pennington2014glove} and Swivel \cite{Shazeer2016} learn embeddings that capture lexical and semantic information. However, the current lack of a standard evaluation practice make it hard to study which specific relations embeddings can effectively capture, nor how to best quantify the signal of such relations.  


At the same time, embeddings as a knowledge representation mechanism is being explored by the traditionally symbolic community that produced semantic networks and knowledge graphs. Algorithms based on \textbf{knowledge graph (KG) embeddings}, like RDF2Vec\cite{Ristoski2016}, ProjE\cite{Shi2016}, TransE \cite{Bordes2013} and HolE\cite{nickel2016HolE}, learn embeddings representing the concepts, words and relations contained in a KG and provide a vector representation of the knowledge that is explicitly described in it. However, such KG embeddings may only encode knowledge that is already represented in the KG. One application of embeddings for this community is \textbf{generic entity-based KG completion and refinement} \cite{riedel2013univSchemas,melo2017relErrorsInKGs,nickel2016relMLForKGs,Paulheim2017KGRefinement,lin2016relExtractOverInstances}, which tries to use large text corpora to complete a partial knowledge graph. However, such efforts have focused on encyclopedic (e.g. DBpedia) or domain-specific (family, commerce, finance, law) relations between entities rather than lexical relations. On the opposite direction, KGs are also used to refine vector space representations as in \cite{Faruqui2015}. Finally, efforts trying to learn \textbf{lexical semantic relations}~\cite{shwartz2016LexNet,gabor2017vecs4semrels,turney2015recognizingLexEntailment,roller2014distribHypernymDetection,fu2014semHierarchiesViaEmbeddings} have started looking at whether word embeddings (and similar distributional approaches) can be used to predict certain types of lexical semantic relations. Many of these approaches suffer from the difficulty to generate a training dataset that serves this purpose~\cite{levy2015doEmbedsLearnRels} and most of them focus on hypernymy relations or lexical inference, still a limited fragment of the whole spectrum of possible lexical semantic relations. Also, WordNet tends to be the only source of evidence used in such work, which may hinder reaching a general understanding of the matter. 

\begin{figure*}[h]
  \includegraphics[width=0.9\textwidth]{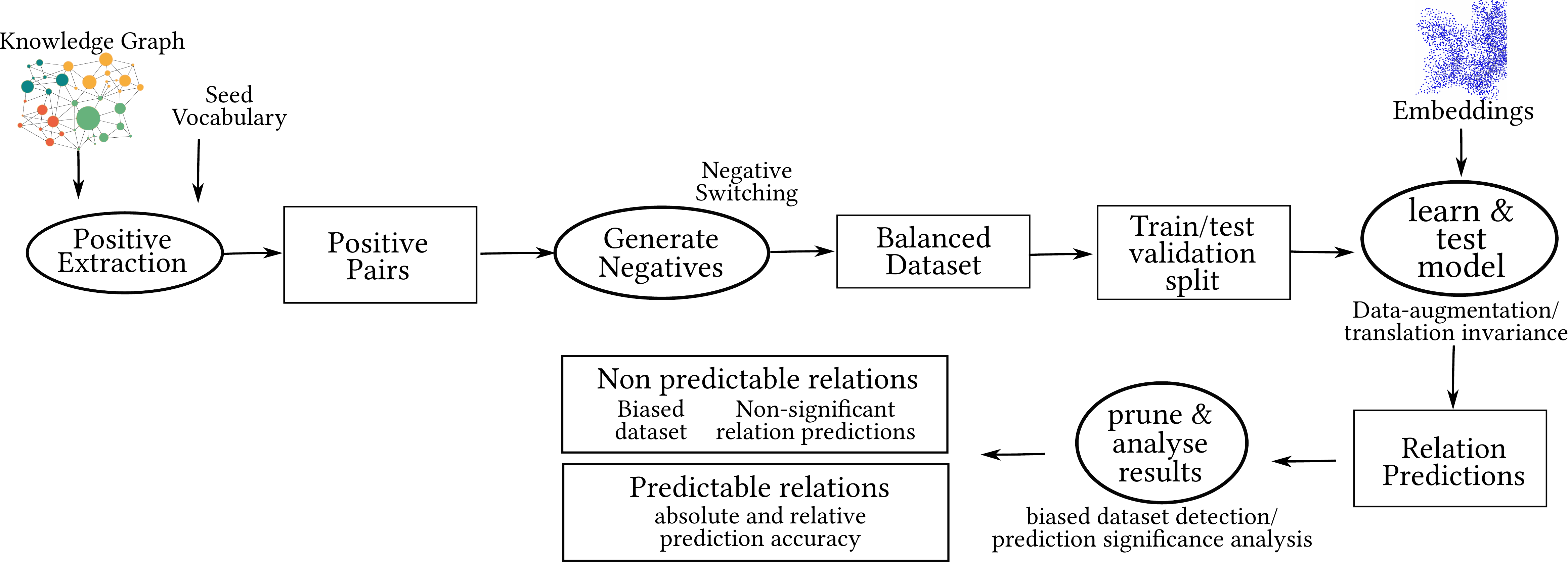}
  \caption{Generic approach for evaluating the predictive power of word and concept embeddings using a knowledge graph as a silver standard.}
 \label{fig:relpredict}
\end{figure*}

In previous work\cite{denaux2017vecsigrafo,denauxvecsigrafo}\footnote{\href{https://bit.ly/2LykQ9T}{Vecsigrafo: Corpus-based Word-Concept Embeddings}} on joint word-concept embeddings, these outperformed word-only and knowledge graph approaches over a selection of 14 benchmarks on semantic similarity and relatedness as well as word-concept and hypernym relation prediction tasks. In doing so, we suggested it may be possible to extract knowledge about lexical semantic relations from (joint word-concept) embeddings. In this paper, we take into account lessons from the various communities discussed above to propose a method for evaluating to what extent pre-trained embeddings contain knowledge about a wide ranging of relations encoded in a KG. Our contributions are: (i) describing a (largely automated) method for (i.a) word/concept relational dataset generation from a KG (i.b) training of machine learning models and (i.c) evaluation of the generated datasets and trained models; (ii) application of this method to analyse to what extent (ii.a) various corpus- and KG-based pre-trained embeddings capture (ii.b) various types of lexico-semantic relational knowledge and (ii.b) what the effect is various factors such as the size of the corpus and the type of dataset (word or concept).

In Section~\ref{sec:genericMethod} we describe a generic method for measuring the predictive power of embeddings for specific relations. In section~\ref{sec:lexicoSemanticStudy} we describe how we apply this methodology to study lexico-semantic relations and in section~\ref{sec:analysis} we analyse the gathered data.

\section{Measuring Relation Predictive Power of Embeddings with a KG}
\label{sec:genericMethod}


In this section we propose a method for studying the predictive power of word and concept embeddings, depicted in Figure~\ref{fig:relpredict}. The goal is to study how much knowledge about relations is encoded in embedding spaces and can be exploited by machine learning models. Note that our goal is \emph{not} to generate the best models for predicting relations (which could be achieved by combining different sources of evidence). The method consists of three main phases: dataset generation, model training and prediction results analysis.

\paragraph{Preliminaries} We define a \textbf{Knowledge Graph} as a tuple $\langle N, R, E \rangle$, where $N = C \cup I \cup W$ is a set of node identifiers, typically referring to concepts $c \in C$, instances $i \in I$ or words $w \in W$ (human readable names for concepts and instances); $R$ is a set of relation types and $E$ is a set of triples of the form $(r\ n_i\ n_j)$ where $r \in R$ and $n_i,n_j \in N$. We define an \textbf{embedding space} $S$ as a tuple $\langle V, F_d \rangle$, where $V$ is a vocabulary, and $F_d: V \longrightarrow \mathbb{R}^d$ is a function that maps elements in the vocabulary to its embedding: a vector of dimension $d$.

The main \emph{inputs} for our method are a KG $k = \langle N_k, R_k, E_k \rangle$ and one or more embedding spaces $s_i \in S'_k \subset S_k \subset S$. $S_k$ is the set of all embedding spaces that have a vocabulary that overlaps with $N_k$. $S'_k$ is a subset of embedding spaces to study. The main \emph{output} of our method is a classification of tuples $(r\ s)$, where $r \in R_k$ and $s \in S'_k$, into \emph{predictable} and \emph{non-predictable}. Furthermore, for each such tuple, we also derive absolute and relative relation prediction metrics, providing a numerical assessment of how well the embedding space $s$ captures or encodes knowledge about relationship $r$.


\paragraph{Dataset generation}
In the first phase, we aim to generate datasets $\delta \in D$, where each dataset $\delta$ is a finite set of triples of the form $\delta_r = \{\langle n_i\ n_j\ l\rangle\}$, where $l$ is a classification label such that $l = 1$ if $(r\ n_i\ n_j) \in E_k$ and $l = 0$ otherwise. We generate datasets between \emph{words} $\delta^w$ with tuples $\langle w_i\ w_j\ l\rangle$, word-concepts $\delta^{wc}$ with tuples $\langle w_i\ c_j\ l\rangle$ and concepts $\delta^c$ with tuples $\langle c_i\ c_j\ l\rangle$.

As a first step, we select a \textbf{seed vocabulary} $V_{\mathrm{seed}} = \bigcap_{i=1}^{|S'_k|} V_{s_i}$, as the intersection of all the studied vocabularies.
Next, $V_{\mathrm{seed}}$ and $k$ are used to \textbf{extract a set of positive pairs} for each $r \in R_k$; these correspond to partial datasets $\delta_r^+ = \{\langle n_i\ n_j\  1 \rangle \mid (r\ n_i\ n_j) \in E_k \wedge n_i,n_j \in V_{\emph{seed}}\}$. KGs frequently define relations at a concept level, therefore, this initial dataset will typically only contain concepts. To also generate word-concept datasets for such relations, we also extract partial datasets
 $\{ \langle w_i\ c_j\ 1\rangle \mid (r\ c_i\ c_j) \wedge (r_w\ w_i\ c_i) \}$, where $r_w$ is the word-to-concept relation in $k$. Similarly, we can extract a word dataset for $r$ using  $\{ \langle w_i\ w_j\ 1\rangle \mid (r\ c_i\ c_j) \wedge (r_w\ w_i\ c_i) \wedge (r_w\ w_j\ c_j) \}$.

  Besides generating positive datasets for relations $r \in R_k$, we also generate \emph{random datasets}. These datasets serve as baselines and will be used later on to prune biased datasets. Since the generated datasets $\delta_r^+$ vary in size depending on the number of relation tuples in $E_k$, we generate sets of varying sizes \footnote{In this work we use sizes 200, 500, 1K, 5K, 10K and 50K.} of ``positive'' random pairs $\{\langle n_i\ n_j\ 1\rangle \}$, where $n_i$ and $n_j$ are randomly sampled from $N_k$. We refer to these datasets as $\delta_{\mathrm{rand},x}^+$ where $x$ is the number of positive pairs generated.

 We need both positive and negative examples in the datasets to train a model. For ease of training, we aim to \textbf{generate balanced datasets} with the same number of positive and negative examples. Randomly \textbf{generating negative pairs} based on the seed vocabulary or selecting positive pairs of a different relation type is not optimal because models can learn to identify words/concepts associated to the relation rather than the relation itself~\cite{levy2015doEmbedsLearnRels}. Instead, we apply \emph{negative switching}, whereby positive pairs are switched based on the subject and object vocabularies for the relation, i.e. from positive pairs $\langle n_i\ n_j\ 1\rangle$ and $\langle n_k\ n_l\ 1\rangle$; thus subject vocabulary $\{n_i, n_k\}$ and object vocabulary $\{n_j, n_l\}$, it is possible to generate negative pairs $\langle n_k\ n_j\ 0\rangle$ and $\langle n_i\ n_l\ 0\rangle$. Albeit practical, this approach does not take into account transitive relations and assumes a closed world. Also, for relations with unbalanced subject-object vocabularies, it may be impossible to generate sufficient negative examples; in such cases we fall back to selecting pairs from other relation types or generating random pairs.

 \paragraph{Model training}
 We next use the generated datasets to train binary classification machine learning models for each studied embedding space $s \in S'_k$, we refer to the resulting trained models as $m^{\delta,s,t}$, where $\delta$ is the generated dataset and $t$ is the model type. We do not specify a type of machine-learning model to be used; however, since the input and output are very simple, we expect fully-connected neural networks to be suitable and use these in our experiments.

 Each sample $\langle n_i n_j l \rangle$ in the dataset is converted into an \emph{input vector for the model} by combining the embeddings for the subject and object arguments; i.e. $F_s(n_i) \odot F_s(n_j)$. In this work we use vector concatenation for $\odot$, but other operations are also possible. However, to prevent models from simply learning which embeddings are associated with a relation type, during training we apply \emph{input perturbation}: for each batch, we generate a random vector $v \in \mathbb{R}^d$ and add it to both the subject and object embedding. Thus the final input to the model is $(F_s(n_i) \oplus v) \odot (F_x{n_j} \oplus v)$. This ensures that the difference between the embeddings is the same as for the original pair, but no individual embedding is seen twice. Also, since different embedding spaces have different scales, we adapt the amount of perturbation to each embedding space by scaling the random perturbation to be within the standard deviation of the embedding space. 

 Furthermore, to verify that the generated dataset does not encode information about the relation, we also train models using a baseline random embedding space $s_{\mathrm{rand}} \in S_k$ for the seed vocabulary. Since the embeddings in $s_{\mathrm{rand}}$ are random, they cannot encode any information about $n \in V_{\mathrm{seed}}$ and their relations $r \in R_k$.

 Each generated dataset is split into training, (validation) and testing subsets; the latter is used to evaluate the trained model by calculating metrics: precision, recall, accuracy and f1. Since the performance of a model is affected by random initialization of its parameters, we propose to perform multiple runs, resulting in $m_i^{\delta,s,t}$ for $1 \leq i \leq \#_{\mathrm{runs}}$. This allows us to calculate the average and standard deviation for each of the collected metrics. Below, we use $\mu^\upsilon_{\delta,s}$ to refer to the average for metric $\upsilon$ for models trained on dataset $\delta$ and embedding space $s$. If $s$ is omitted, the average is taken over all models trained on $\delta$. Likewise for the standard deviation $\sigma^\upsilon$.

 \paragraph{Prediction results analysis}
 Once we have trained all our models, we can do some statistic analysis on the collected metrics. Since the dataset generation and model training are mostly automated, our main goal here is to identify any non-predictable relations. Using our baselines, we can discard results based on two reasons: (i) the generated dataset is biased and (ii) the learned model's results are not significant.

 First, we define \textbf{baseline ranges for prediction metrics}. For this, we use the metrics gathered for the generated random datasets $\delta_{\mathrm{rand,x}}$ and define the range thresholds as $\tau^{\upsilon_\mathrm{min}}_{\mathrm{biased}} = \mu^\upsilon_{\delta_{\mathtt{rand},x}} - 2 \sigma^\upsilon_{\delta{\mathrm{rand},x}}$ and $\tau^{\upsilon_\mathrm{max}}_{\mathrm{biased}} = \mu^\upsilon_{\delta_{\mathtt{rand},x}} + 2 \sigma^\upsilon_{\delta{\mathrm{rand},x}}$. Any metrics within these ranges could be due to chance with 95\% probability.

 We consider a dataset to be \textbf{biased} if models can perform well on them regardless of whether the embeddings used to train the model encode any information. Intuitively, these are datasets which are imbalanced in some way allowing the model to exploit this imbalance during prediction, but that do not reflect the knowledge encoded in the embeddings. To detect these, we look at model results for models trained on random embeddings (i.e. on models $m^{\delta_r,s_{\mathrm{rand}},t}$). We say that $\delta_r$ is biased if $\mu^\mathrm{f1}_{\delta_r, s_{\mathrm{rand}}}$ is outside of the  $[ \tau^{\mathrm{f1}_\mathrm{min}}_{\mathrm{biased}}, \tau^{\mathrm{f1}_\mathrm{max}}_{\mathrm{biased}} ]$ range. The rationale is that even with random embeddings, such models were able to perform outside of the 95\% baseline ranges.

 We consider a trained model $m^{\delta_r,s,t}$ to be \textbf{significant} if its predictions are statistically better than predictions made by $m^{\delta_r,s_\mathrm{rand},t}$. Intuitively, this indicates that the embedding space $s$ contributes information about relation $r$ with high probability. Formally, we say that $m^{\delta_r,s,t}$ is significant if $\mu^{\mathrm{f1}}_{\delta_r,s} - \mu^{\mathrm{f1}}_{\delta_r, s_{\mathrm{rand}}} > 2\mathrm{max}(\sigma^{\mathrm{f1}}_{\delta_r,s}, \sigma^{\mathrm{f1}}_{\delta_r, s_{\mathrm{rand}}})$.


\section{Measuring Lexico-Semantic Knowledge in Embeddings}
\label{sec:lexicoSemanticStudy}

We applied the method described above to two lexico-semantic KGs (WordNet and Sensigrafo) and several embedding spaces derived from different corpora and embedding algorithms. In this section we describe how we applied the methodology and present a summary of the results obtained. In the next section we discuss analyze the results.\footnote{The code we used is available at \url{https://github.com/rdenaux/embrelassess}}


\subsection{Knowedge Graphs and Relations}
\label{sec:KGsAndRels}

We applied our methodology to two KGs: (i) WordNet, a well known lexico-semantic semantic network and (ii) Sensigrafo, a proprietary lexical knowledge graph developed by Expert System. Although similar in structure and scope --both KGs aim to provide a sense lexicon per language and use hypernymy as the main relation between senses--, Sensigrafo has been developed independently and is tightly coupled to Expert System's text analysis pipeline, enabling state of the art word sense disambiguation with claimed 90\% accuracy.

\begin{table}
  \centering
  \small
  \begin{tabular}{lll}
 & Sensigrafo & WordNet\\
\hline
version & 14.2 & 3.0\\
words/lemmas & 400K & 155K\\
syn(set/con) & 300K & 118K\\
relations & 55 & 27\\
derived rel datasets & 149 & 27\\
pair types & lem2(lem,syn,POS) & lem2lem\\
 & syn2(syn,POS) & \\
\hline
  \end{tabular}
  \caption{Lexical Knowledge Graphs used.}\label{tab:kgs}
\end{table}

Table~\ref{tab:kgs} provides an overview and comparison for the KGs. WordNet does not have an explicit identifier for concepts; instead, it defines sets of synonyms, i.e. lemmas with a shared sense. Sensigrafo calls such sets syncons (synonym-concepts), since they refer to specific concepts and assigns a unique identifier to each. There are differences in terms of size, granularity of concepts, structure of the network (e.g. choice of central concepts) and types of relations. Finally, because Sensigrafo is developed and maintained as part of a text analytics pipeline, some of its features are tailored and biased towards supporting  functionality and domains required by Expert System customers. By comparison, being a community effort WordNet may benefit from a wider set of stakeholders.

\begin{table*}
  \centering
  \small
  \begin{tabular}{llllrrrrr}
type & name & KG & example pair & lem pairs & syn  pairs & obj:subj & voc tot\\
\hline
Lexical & also see & w & mild-temperate & 5800 &  & 1.25 & 2339\\
 & derivation & w & revoke-revocation & 118888 &  & 1.0 & 18094\\
 & pertainym & w & regretfully-sorry & 6516 &  & 1.37 & 6079\\
     & participle of & w & operating-operate & 81 &  & 2.13 & 94\\
   \noalign{\smallskip}
Hypernymy & hypernym* & w & cinnamon-spice & 110650 &  & 0.45 & 27048\\
 & sup/subnomen & s & ditto & 56413 & 33696 & 1.98 & 19768\\
 & super/subverbum & s & kick-move & 37660 & 9426 & 1.81 & 6630\\
 & instance hypernym* & w & gemini-constellation & 2358 &  & 0.44 & 1526\\
   \noalign{\smallskip}
Categorical & category domain* & w & fly-air, tort-law & 9116 &  & 0.13 & 4166\\
 & sensiDomain & s & antitrust case-commercial law & 28610 &  & 0.02 & 17636\\
 & usage domain* & w & squeeze-slang & 846 &  & 0.09 & 395\\
 & region domain* & w & legionnaire-france & 1349 &  & 0.12 & 546\\
 & noun cat & s & flora-natural object &  & 186797 & 0.004 & 50257\\
 & verb cat & s & belong-v. of generic state &  & 27388 & 0.01 & 11498\\
 & tag & s & Christmas Eve-calendar day &  & 31775 & 0.01 & 16825\\
   \noalign{\smallskip}
Meronymy & member meronym* & w & archipelago-island & 1315 &  & 1.51 & 1147\\
 & substance meronym* & w & brine-sodium & 369 &  & 0.98 & 378\\
 & part meronym* & w & aeroplane-wing & 6403 &  & 1.39 & 4054\\
 & omni/parsnomen & s & construction-roofing & 807 & 3110 & 1.22 & 597\\
   \noalign{\smallskip}
Synonymy & synonym & w & encourage-promote & 74822 &  & 1.0 & 22304\\
 & synonym & s & ditto & 69974 &  & 1.0 & 19130\\
   \noalign{\smallskip}
Concept Simil & similar & w & big-immense & 20464 &  & 1.0 & 6790\\
 & attribute & w & short-length, good-quality & 1718 &  & 1.0 & 859\\
 & cause & w & secure-fasten & 719 &  & 0.93 & 445\\
 & syncon-cause & s & ring-sound, fright-fear & 584 &  & 0.87 & 458\\
 & entailment & w & look-see, peak-go up & 1519 &  & 0.89 & 956\\
 & syncon-implication & s & overtake-compete & 1358 & 291 & 0.83 & 870\\
 & verb group & w & shift-change, keep-prevent & 4944 &  & 1.0 & 1193\\
 & syncon-corpus & s & find-strike & 97707 & 39644 & 1.11 & 10502\\
 & syncon-unification & s & ritual-rite, enclose-envelop & 6599 & 2392 & 0.99 & 3366\\
 & antonym & w & release-detain & 9310 &  & 1.0 & 4651\\
 & (adj,n,v)antonym & s & ditto & 4656 & 1728 & 1.03 & 1823\\
   \noalign{\smallskip}
Positional & s-adjective-class & s & nightmarish-account & 131853 & 33037 & 0.69 & 13972\\
 & adverb-(n,v,adj,adv) & s & below-criteria & 15598 & 3367 & 1.12 & 2509\\
 & verb-object & s & counter-illness & 108046 & 23163 & 1.24 & 9087\\
 & verb-subject & s & less-tension, plunge-index & 46949 & 10131 & 1.01 & 7126\\
   \noalign{\smallskip}
Prepositional & 12 internoun & s & meeting-colleague &  &  & 0.83 & 10748\\
 & 10 verb prep noun & s & break down-tear &  &  & 0.56 & 1848\\
 & verb prep verb & s & set-rise & 287 &  & 1.19 & 196\\
 & 4 adj prep noun & s & eligible-admission &  &  & 1.62 & 299\\
   \noalign{\smallskip}
Geographic & geography & s & Brussels-Brussels Capit. Reg. &  & 1555 & 0.23 & 1584\\
   \noalign{\smallskip}
Part-of-Speech & Noun & s & entity-GNoun &  & 45961 & n/a & 76138\\
 & ProperNoun & s & Underground Railway-GPNoun &  & 4134 & n/a & 8273\\
 & Verb & s & cleanse-GVerb &  & 11343 & n/a & 22691\\
 & Adjective & s & record-GAdjective &  & 12585 & n/a & 25175\\
 & Adverb & s & in my opinion-GAdverb &  & 2110 & n/a & 4225\\
   \end{tabular}
  \caption{Overview of lexico-semantic relations studied.}
  \label{tab:relsA}
\end{table*}

WordNet and Sensigrafo provide 27 and 55 types of relations respectively which we used to generate datasets. Table~\ref{tab:relsA} shows an overview of the relations extracted; since discussing the 88 relations individually is unwieldy, we group the relations into types following --and expanding-- a categorization described in the WordNet manual\footnote{\url{https://wordnet.princeton.edu/documentation/wninput5wn}}. It suggests a top-level distinction between \textbf{Lexical} --those that hold between words-- and \textbf{Semantic} --those that hold between concepts-- relations and lists relations \emph{also see}, \emph{antonym}, \emph{derivation}, \emph{participle} and \emph{pertainym} as lexical. However, \emph{antonym} is clearly based on the word's meaning. \emph{Derivation} and \emph{participle} seem to be truly lexical, while the others can also contain pairs which are semantically related, as shown by the example pairs in the table. Sensigrafo does not contain purely lexical relations.

The main relation type used by both KGs is \textbf{hypernymy}, which relates narrower to broader concepts (or instances to concepts). Although hypernymy is transitive, we only consider direct relations explicitly stated in the KG during the dataset generation.

\textbf{Categorical relations} are those that associate a concept with some category. When categories are concepts, they can be seen as an indirect hypernymy. WordNet has three types of categories called domains: \emph{category} --concept to a topic domain--, \emph{usage} --concepts to a type of use-- and \emph{region} --concepts to places. Sensigrafo has a set of core noun and verb concepts --called noun or verb categories, as well as tags--  which form the backbone of the hypernymy hierarchy. These category syncons tend to be quite abstract concepts, hence we have only considered these relations at the syn2syn level, as they do not have lemmas in the seed vocabulary. Sensigrafo also defines a list of about 400 domains (which are not syncons), which is similar to WordNet's \emph{category domain}.

\textbf{Meronymy} relates concepts in whole-member relations. WordNet distinguishes between \emph{membership} --part of group--, \emph{substance} --when something is made of substances that are orders of magnitude smaller than the whole-- and \emph{part} -- remaining cases. Sensigrafo does not distinguish between these cases and uses this type of relation sparingly.

\textbf{Synonymy} can be defined between lemmas in the same synset/-con and is a special case of \textbf{conceptual similarity}. The \emph{similarity} relation in WordNet captures pairs of similar adjectives. The \emph{attribute} relation in WordNet relates adjectives describing a value for a noun (similar, but more limited than the \emph{adjective-class} relation in Sensigrafo). The \emph{cause} relation in WordNet describes causality between verbs and is similar in Sensigrafo. The \emph{entailment} relation in WordNet describes an entailment between verbs, similarly to the \emph{syncon-implication} in Sensigrafo, which can also be applied to non-verbs. \emph{Verb group} in WordNet groups similar verbs. \emph{Syncon-unification} is assigned by linguists to syncon pairs that could be merged as a single syncon. \emph{Antonym} describes concepts with opposite meanings.

\textbf{Positional} relations encode co-locations of concepts and are only provided by Sensigrafo. Two main subtypes: one relates adjectives or adverbs to other concepts while the second relates verbs to concepts that appear as the verb subject or object.

Sensigrafo provides about two dozens of \textbf{prepositional} relations between concepts. Such relations are of the type POS+preposition-POS. E.g. the pair \emph{rival-titleholder} has relation \emph{noun+to-noun}. Sensigrafo also encodes \textbf{geographic} relations between places. However, since the seed vocabulary did not contain many place names, we could not generate a dataset between lemmas.

Finally, we also generated datasets relating syncons to their \textbf{part-of-speech} (POS) besides the various $\delta_{\mathrm{random},x}$ as explained in Section~\ref{sec:genericMethod} (not included in Table~\ref{tab:relsA} ).




We generated a total of 176 datasets, based on the 88 relations and a seed vocabulary consisting of 76K concepts and 71K lemmas (roughly corresponding to our smallest embedding space, which was trained on a disambiguated version of the English United Nations corpus~\cite{ziemski2016united}). For WordNet, we only generated datasets between lemmas. For Sensigrafo we also generated datasets between words, word/concepts and concepts.

\subsection{Embeddings and Corpora}

The word and concept embeddings studied were derived from 6 algorithms. Three provided embeddings based on sequences of either words or syncons: GloVe~\cite{pennington2014glove}, FastText~\cite{Bojanowski2017}, Swivel~\cite{Shazeer2016}. Two provided joint word and concept embeddings: Vecsigrafo~\cite{denaux2017vecsigrafo} based on a disambiguated corpus and HolE~\cite{nickel2016HolE}, which directly generates embeddings from a KG (not from a corpus) and thus serves as a reference point. The embeddings were either publicly available or provided to us by \cite{denaux2017vecsigrafo}. We also generated random embeddings $s_{\mathrm{rand}}$ for the seed vocabulary.

The embeddings were trained on three different corpora, which we chose to study whether relation prediction capacity varies depending on the corpus size: the English United Nations corpus\cite{ziemski2016united} (517M tokens), the English Wikipedia (just under 3B tokens) and Common Crawl (around 840B tokens). Although we aimed at using embeddings with 300 dimensions, in a few cases we had to diverge as the embeddings were only available with other dimensions.

\subsection{Training and Results}
We generated models based on fully connected neural networks (NN) with either 2 or 3 hidden layers. Initially we also generated models with logistic regression, which consistently underperformed. The 2 and 3 layer NNs generally produced similar results, suggesting they converge for the given datasets. For the standard case of embedding dimension 300, the input to the net is a vector of 600 dimensions, followed by hidden layers of 750 and 400 nodes for the NN2; and (750, 500, 250) for the NN3 (similar architectures were defined for single embedding relations such as syn2POS). The output layer has 2 nodes and uses 1-hot-encoding to encode positive or negative examples. For regularization, we use dropout between layers with value 0.5. We used a heuristic rule that varies the number of epochs to train a dataset depending on its size. Datasets with $<300$ positive examples are trained for 48 epochs, those with $<5K$ for 24, $<30K$ for 12 and large datasets are trained for 6 epochs. We use the Adam optimizer with learning rate $1^{-5}$ and the cross entropy loss function. A scheduler reduces the learning rate on plateau. All of these hyper-parameters where derived through trial and error with a few sample relation datasets and kept constant to automatically train models without manual inspection. We used a random 90, 5, 5 split for training, validation and test from the input dataset. For WordNet relations we used a mixture of NN2 and NN3 models and trained each model either 3 or 5 times. As training the models is the main bottleneck of our approach, for Sensigrafo relations we only trained NN3 models, training each model 3 times.

We trained a total of 10,560 models, resulting in 1,596 evaluation metrics averaged over $n$ runs: 126 for the random relation datasets, 149 trained on random embeddings, 1,065 for Sensigrafo relations and 268 for WordNet relations. Each run resulted in metrics for relation prediction on unseen pairs during training. 

\section{Analysis and Discussion of Results}
\label{sec:analysis}


\begin{table*}
  \centering
  \small
  \begin{tabular}{ll @{\hskip 0.25in}rr @{\hspace*{5mm}}rrr@{\hskip 0.2in}rr @{\hskip 0.2in}rr}
    \hline
    &  & \multicolumn{2}{c}{}   & \multicolumn{5}{c}{models} & \multicolumn{2}{c}{metrics}\\ 
    \cline{5-9}
    dataset rel & KG & \multicolumn{2}{c}{datasets}   & & \multicolumn{2}{c}{non-predictable} & \multicolumn{2}{c}{``predictable''} & absolute & relative \\
    \cline{3-4}\cline{6-7}
 &  & \# & \# biased & \# & \% biased & \% not signif. & \textbf{\% better} & \% worse & $\mu_{\mathrm{f1}}$ & $\Delta \mu_{\mathrm{f1}}$\\
\hline
all & both & 156 & 38 & 1281 & \textbf{28.6} & \textbf{46.5} & \textbf{23.9} & \textbf{0.9} & 0.687 & 0.173 \\[3pt]
all & wn & 19 & 6 & 216 & 27.8 & 39.4 & 30.1 & 2.8 & 0.684 & 0.149\\
all & sensi & 137 & 32 & 1065 & 28.8 & 48. & 22.6 & 0.6 & 0.688 & 0.181\\[4pt]
concept & sensi & 44 & 10 & 293 & 22.9 & \textbf{73.7} & \textbf{3.4} & 0. & \textbf{0.870} & \textbf{0.608}\\
word/concept & sensi & 43 & 3 & 172 & \textbf{7.} & 43.6 & \textbf{48.8} & 0.6 & 0.664 & 0.162\\
word & sensi & 50 & 19 & 600 & \textbf{38.} & 36.7 & 24.5 & 0.8 & 0.690 & 0.162\\[4pt]
lexical\(_{\text{w}}\) & wn & 4 & 0 & 32 & 0. & \textbf{75.} & 25. & 0. & 0.652 & 0.167\\[2pt]
hypernym\(_{\text{c}}\) & sensi & 2 & 1 & 14 & 50. & 42.9 & 7.1 & 0. & \textbf{0.916} & \textbf{0.696}\\
hypernym\(_{\text{w/c}}\) & sensi & 2 & 1 & 8 & 50. & 0. & \emph{50.} & 0. & 0.699 & 0.148\\
hypernym\(_{\text{w}}\) & sensi & 2 & 1 & 24 & 50. & 0. & \emph{50.} & 0. & 0.713 & 0.143\\
hypernym\(_{\text{w}}\) & wn & 2 & 1 & 24 & 50. & 4.2 & 37.5 & \textbf{8.3} & \textbf{0.756} & 0.106\\[2pt]
    
categ\(_{\text{c}}\) & sensi & 3 & 1 & 21 & 33.3 & 52.4 & 14.3 & 0. & \textbf{0.891} & \textbf{0.671}\\
categ\(_{\text{w/c}}\) & sensi & 4 & 1 & 16 & 25. & 12.5 & \textbf{62.5} & 0. & 0.744 & 0.263\\
categ\(_{\text{w}}\) & sensi & 1 & 1 & 12 & \textbf{100.} & 0. & 0. & 0. &  & \\
categ\(_{\text{w}}\) & wn & 4 & 4 & 40 & \textbf{100.} & 0. & 0. & 0. &  & \\[2pt]
    
meronym\(_{\text{c}}\) & sensi & 2 & 0 & 14 & 0. & \textbf{92.9} & 7.1 & 0. & 0.667 & 0.020\\
meronym\(_{\text{w/c}}\) & sensi & 1 & 0 & 4 & 0. & \textbf{50.} & \emph{50.} & 0. & 0.664 & 0.296\\
meronym\(_{\text{w}}\) & sensi & 1 & 0 & 12 & 0. & \textbf{91.7} & 0. & \textbf{8.3} &  & \\
meronym\(_{\text{w}}\) & wn & 3 & 0 & 48 & 0. & \textbf{66.7} & 31.2 & 2.1 & 0.708 & 0.166\\[2pt]

synon\(_{\text{c}}\) & sensi & 0 & 0 & 0 & 0. & 0. & 0. & 0.&  & \\
synon\(_{\text{w/c}}\) & sensi & 1 & 0 & 4 & 0. & 25. & \textbf{75.} & 0. & \textbf{0.804} & 0.249\\
synon\(_{\text{w}}\) & sensi & 1 & 0 & 12 & 0. & 8.3 & \textbf{91.7} & 0.  & 0.677 & 0.114\\
synon\(_{\text{w}}\) & wn & 1 & 0 & 8 & 0. & 0. & \emph{100.} & 0. & 0.680 & 0.135\\[2pt]
    
simil\(_{\text{c}}\) & sensi & 6 & 1 & 42 & 16.7 & 81. & 2.4 & 0. & \textbf{0.909} & \textbf{0.688}\\
simil\(_{\text{w/c}}\) & sensi & 7 & 0 & 28 & 0. & 57.1 & 42.9 & 0.  & 0.624 & 0.210 \\
simil\(_{\text{w}}\) & sensi & 7 & 1 & 84 & 14.3 & 38.1 & \textbf{47.6} & 0. & 0.628 & 0.128\\
simil\(_{\text{w}}\) & wn & 5 & 1 & 64 & 12.5 & 43.8 & 39.1 & \textbf{4.7} & 0.655 & 0.153\\[2pt]

position\(_{\text{c}}\) & sensi & 6 & 1 & 42 & 16.7 & \textbf{78.6} & 4.8 & 0. & \textbf{0.886} & 0.667\\
position\(_{\text{w/c}}\) & sensi & 6 & 1 & 24 & 16.7 & 37.5 & \textbf{45.8} & 0. & 0.721 & 0.120\\
position\(_{\text{w}}\) & sensi & 7 & 5 & 84 & \textbf{71.4} & 21.4 & 6. & 1.2 & 0.703 & 0.069\\[2pt]
    
prepos\(_{\text{c}}\) & sensi & 19 & 4 & 133 & 21.1 & \textbf{78.9} & 0. & 0. &  & \\
prepos\(_{\text{w/c}}\) & sensi & 22 & 0 & 88 & 0. & 51.1 & \textbf{47.7} & 1.1  & 0.629 & 0.124\\
prepos\(_{\text{w}}\) & sensi & 27 & 11 & 324 & 40.7 & 48.5 & 9.9 & 0.9 & 0.677 & 0.097\\[2pt]
    
    
POS\(_{\text{c}}\) & sensi & 5 & 1 & 20 & 20. & \textbf{70.} & 10. & 0. & \textbf{0.882} & \textbf{0.665}\\
POS\(_{\text{w/c}}\) & sensi & 0 & 0 & 0 & 0. & 0. & 0. & 0. &  & \\
POS\(_{\text{w}}\) & sensi & 4 & 0 & 48 & 0. & 2.1 & \textbf{97.9} & 0. & \textbf{0.746} & 0.262\\
\hline
\end{tabular}
  \caption{Overview of results.}
  \label{tab:resultsOverview}
\end{table*}

\paragraph{Biased datasets and non-significant models}
We apply the prediction result analysis  described in Section~\ref{sec:genericMethod}: based on 12 random datasets $\delta_{\mathrm{rand},x}$ and 126 prediction results we obtained $\mu^{\mathrm{f1}}_{\delta_{\mathrm{rand}}}=0.41$ and  $\sigma^{\mathrm{f1}}_{\delta_{\mathrm{rand}}}=0.12$, resulting in a baseline range for prediction metrics (i.e. $[ \tau^{\mathrm{f1}_\mathrm{min}}_{\mathrm{biased}}, \tau^{\mathrm{f1}_\mathrm{max}}_{\mathrm{biased}} ]$) of $[0.16, 0.65]$.

Using this range, we identify 38 of 156 datasets as being \emph{biased}. With other words, even though we took care not to generate non-biased datasets by using negative switching, a little more than a quarter of the datasets generated in this way contains clues about the relation. Interestingly, word-pair datasets are much more likely to be biased (38\%) while word/concept pairs are unlikely to so (only 7\%) (see Table~\ref{tab:resultsOverview}). These result suggest that using KGs as a \emph{silver standard}~\cite{Paulheim2017KGRefinement} is of limited use for word-pair prediction, but is suitable when linking words to concepts.

We see that using a KG to build datasets for certain kind of relations is very difficult. This is the case for word pair datasets for categorical relations (all of the 52 generated datasets were biased); this is likely due to such relations being highly unbalanced with only a few words in the object position, further compounded by ambiguity of words compared to concepts. We think word ambiguity also plays a role in the difficulty producing non-biased datasets for positional relations.

On the bright side, about 71\% of the models were trained on a non-biased dataset and we use these to identify (non)significant models. Overall, we found that 46.5\% of the trained models using pre-trained embeddings were not statistically significant different from a baseline using random embeddings. About 1\% of the models performed worse than the baseline. These were typically small datasets and relations that were hard to learn. In any case this 1\% is below the 2.5\% that can be expected by using the $2\sigma$ threshold for significance.

Conversely, only about 24\% of the models trained using pre-trained embeddings significantly outperformed the baseline. For word/concept pair datasets this percentage jumps to almost 49\% while for concept pairs it plummets to only 3.4\%. Since most of the pre-trained embeddings we are using are corpus-based, this shows that such embeddings have trouble capturing the relations at the purely conceptual level. At the same time, these models are relatively good at relating words to concepts using semantic relations. In the sections below we only discuss results for the models that outperformed their baseline.

\subsection{Relation Types in Embeddings}

In the last two columns of Table~\ref{tab:resultsOverview} we see the absolute and relative f1 measures for the models that significantly outperformed the random baseline. We see that most of the results with an average f1 score higher than 0.8 are for relations between concepts; these were all achieved by training models on the HolE embeddings. Swivel embeddings also obtained good results for predicting categorical relation (0.846 f1, but this result just cleared the $2\sigma$ significance threshold). One model trained on FastText embeddings achieved (0.666 f1 on a meronymy relation). This confirms that \emph{the studied corpus-based embeddings are not capable of capturing semantic relations at the concept level}.

For the models trained on word/concept or word pairs, we can go through the relation types. Absolute prediction accuracy for \textbf{lexical relations} is poor, the Vecsigrafo and GloVe are the only embeddings that can predict pertainym relations with over 0.7 f1 score. Prediction of \textbf{hypernym} relations is good; the best performing model is trained on HolE on a word/concept dataset (0.797 f1); from the corpus-based embeddings, FastText performs best (0.766). Prediction of \textbf{categorical} relations (only for word/concept pairs) is quite good; corpus-based embeddings using Vecsigrafo perform similarly to HolE (around 0.82 f1) for nouns; corpus-based embeddings have trouble with categorical relations between verbs, with performance dropping below 0.6 compared to 0.8 for HolE. For \textbf{meronymy} relations, the datasets generated on Sensigrafo were rather small, resulting in f1 performances around 0.69 with Vecsigrafo embeddings; WordNet-derived datasets produce good performance with FastText, GloVe and Vecsigrafo having scores between 0.75 and 0.81 for the part-meronym relation; substance- and member-meronym relations accuracy drops below 0.7. Performance for \textbf{synonym} relations depends strongly on the used embedding; the best performers are HolE (0.93 f1), Vecsigrafo (0.79) and FastText (0.75) while Swivel performs poorly. Prediction of \textbf{similarity} relations is mediocre; the best embeddings are HolE (0.81), followed by GloVe and FastText (around 0.74); depending on the relation, prediction f1 can drop to 0.6. For \textbf{positional} relations, Vecsigrafo (0.76) outperforms even HolE (0.74) with FastText and Glove trailing (0.7); this is similar for \textbf{prepositional} relations. For \textbf{part-of-speech}, FastText and GloVe perform well (f1 above 0.8).

\subsection{Impact of embedding type and corpus size}

Unsurprisingly, the larger the corpus the better the overall results for Sensigrafo datasets. Average f1 scores for embeddings trained on the UN corpus, Wikipedia and Common Crawl were 0.66, 0.69 and 0.73. For WordNet datasets (only lem2lem), the scores are similar: 0.68, 0.72 and 0.71. Thus, although increasing the corpus size helps, for many types of relations, the gain from training on a very large corpus is relatively small.

Table~\ref{tab:algoByPairTypes} summarizes the performance of the different embedding learning algorithms for predicting relations grouped by the pair type. For word pair prediction, FastText outperforms other embedding learning algorithms, including HolE (although this difference is not major). HolE excels at predicting relations between senses, but its performance decreases as lemmas are introduced, since it cannot disambiguate between the senses. Vecsigrafo and GloVe both are not far behind the performance of FastText and HolE, but produce significant predictions for more relations than FastText and HolE. Standard Swivel with words lags behind, especially in the number of relations that it can predict.

\begin{table}
  \centering
  \small
\begin{tabular}{rlrrl}
datasets & algo & F1\(_{\text{avg}}\) & F1\(_{\text{std}}\) & pair type\\
\hline
\textbf{8} & HolE & \textbf{0.90} & 0.02 & concept\\
1 & Swivel & 0.85 & 0.0 & concept\\
1 & FastText & 0.67 & 0.0 & concept\\
  \noalign{\smallskip}
26 & HolE & \textbf{0.67} & 0.09 & word/concept\\
48 & Vecsigrafo & 0.64 & 0.08 & word/concept\\
  \noalign{\smallskip}
38 & FastText & \textbf{0.74} & 0.08 & word\\
12 & HolE & 0.72 & 0.09 & word\\
43 & Vecsigrafo & 0.68 & 0.07 & word\\
43 & GloVe & 0.68 & 0.08 & word\\
21 & Swivel & 0.66 & 0.06 & word\\
\end{tabular}
  \caption{Average F1 scores for predicting relations with different pair types and embeddings.}\label{tab:algoByPairTypes}
\end{table}

\subsubsection{Impact of joint word-concept learning}
Vecsigrafo co-trains word and concept embeddings. Table~\ref{tab:algoByPairTypes} shows that compared to Swivel, this co-training improves, for word pairs, both the number of relations that can be predicted (double the number) and the average f1 score (we assume that the additional predicted relations push the average score down). Similarly, we see that compared to word pairs, Vecsigrafo is able to produce predictions for more relations when considering word/concept pairs. As discussed above, corpus-based, joint word-concept training does not seem to capture relations between at the concept level, suggesting there is room to improve such algorithms.

\section{Conclusion}
This paper presented a methodology for studying whether embeddings capture relations as well as KGs and applied it to study lexico-semantic relations between words and concepts. The results show that for a few relations, word embeddings can outperform embeddings derived directly from KGs. Also, corpus-based embeddings fail to capture relations at the concept level. However, for most relation types, embeddings only can predict relations with an accuracy under 0.7. Our results provide evidence that correct capture of relations should happen at the concept level and may not be achievable with high accuracy at the word level. As future work, we want to apply our method to study contextual embeddings.

\section*{Acknowledgments}
The research reported in this paper is supported by the EU Horizon 2020 programme, under grants European Language Grid-825627 and Co-inform-770302.

\bibliographystyle{named}
\bibliography{lex_sem_rel}





\end{document}